\documentclass[11pt,a4paper]{article}
\usepackage[hyperref]{acl2019}
\usepackage{times}
\usepackage{latexsym}

\usepackage{url}
\usepackage{pbox}
\usepackage[utf8]{inputenc}
\usepackage{fourier} 
\usepackage{array}
\usepackage{makecell}
\usepackage{amsmath}

\aclfinalcopy 

\title{WriterForcing: Generating more interesting story endings}

\author{Prakhar Gupta\thanks{\quad equal contribution} \qquad
  Vinayshekhar Bannihatti Kumar\footnotemark[1]  \qquad
  Mukul Bhutani\footnotemark[1] \qquad
  Alan W Black \\
  School of Computer Science\\
Carnegie Mellon University\\
Pittsburgh, PA\\
  \texttt{\{prakharg, vbkumar, mbhutani, awb\}@cs.cmu.edu} \\
  }

\date{}

\begin{document}
\maketitle
\begin{abstract}
We study the problem of generating interesting endings for stories. Neural generative models have shown promising results for various text generation problems. Sequence to Sequence (Seq2Seq) models are typically trained to generate a single output sequence for a given input sequence. However, in the context of a story, multiple endings are possible. Seq2Seq models tend to ignore the context and generate generic and dull responses. Very few works have studied generating diverse and interesting story endings for a given story context. In this paper, we propose models which generate more diverse and interesting outputs by 1) training models to focus attention on important keyphrases of the story, and 2) promoting generation of non-generic words. We show that the combination of the two leads to more diverse and interesting endings.
\end{abstract}

\section{Introduction}

Story ending generation is the task of generating an ending sentence of a story given a story context. A story context is a sequence of sentences connecting characters and events. This task is challenging as it requires modelling the characters, events and objects in the context, and then generating a coherent and sensible ending based on them. Generalizing the semantics of the events and entities and their relationships across stories is a non-trivial task. Even harder challenge is to generate stories which are non-trivial and interesting. In this work, we focus on the story ending generation task, where given a story context - a sequence of sentences from a story, the model has to generate the last sentence of the story.

\begin{table}[h]
\begin{center}
\begin{tabular}{|l|}
 \hline
\textbf{Context} \\ My friends and I did not know what to do \\ for our friends birthday. We sat around the \\ living room trying to figure out what to do. \\ We finally decided to go to the movies. We all \\ drove to the theatre and bought tickets. \\ \\
\hline
\\
\textbf{Specific response (ground truth)} \\ The movie turned out to be terrible but our \\ friend had a good time. \\
\textbf{Generic response (seq2seq output)} \\ We were so happy to see that we had a good  \\ time. \\
\hline
\end{tabular}
\end{center}
\caption{\label{table:dataset-example} For a given story context there can be multiple possible responses. The ones which are more specific to the story are more appealing to the users.}
\end{table}
 
Seq2seq models have been widely used for the purpose of text generation. Despite the popularity of the Seq2seq models, in story generation tasks, they suffer from a well known issue of generating frequent phrases and generic outputs. These models when trained with Maximum Likelihood Estimate (MLE), learn to generate sequences close to the ground truth sequences. However, in story generation tasks, there can be multiple possible reasonable outputs for a given input story context. MLE objective in these models results in outputs which are safe (that is more likely to be present in any output), but also bland and generic. Some examples of such generic outputs in story ending generation task are - "He was sad", "They had a great time", etc. Table~\ref{table:dataset-example} shows an example story from the ROC stories corpus \cite{mostafazadeh2017lsdsem} and the corresponding specific and generic responses. 


There have been many attempts to solve the issue of generation of generic responses. They can be broadly categorized into two categories:

\begin{enumerate}
    \item Use conditionals such as emotions, sentiments, keywords, etc. that work as factors to condition the output on \cite{li-etal-2016-persona, pmlr-v70-hu17e}. When the models focus on these conditionals given as additional input features, they tend to generate outputs which are more relevant and specific to the conditionals, which leads to less generic outputs. In our models, we use the keyphrases present in the story context as conditionals.
    \item Modify the decoding procedure \cite{DBLP:conf/aaai/VijayakumarCSSL18}  with beam search or other variants, or the loss of the decoder \cite{baheti-etal-2018-generating, li-etal-2016-diversity} to encourage the model to generate more diverse outputs.
    Our proposed model uses the ITF loss function suggested by \citet{nakamura2018another} to encourage the decoder to produce more interesting outputs.
\end{enumerate}

We show that our proposed models can generate diverse and interesting outputs by conditioning on the keyphrases present in the story context and incorporating a modified training objective. 
Apart from human judgement based evaluation, we measure performance of the models in terms of 1) Diversity using DISTINCT-1,2,3 metrics and 2) Relevance by introducing an automatic metric based on Story Cloze loss. Experiments show that our models score higher than current state of the art models in terms of both diversity and relevance. 

For reproducibility purposes we are making our codebase open source \footnote{https://github.com/witerforcing/WriterForcing}. 


\section{Related Work}
There has been a surge in recent years to tackle the problem of story generation. One common theme is to employ the advances in deep learning for the task. \citet{jain2017story} use Seq2Seq models \cite{sutskever2014sequence} to generate stories from descriptions of images. \citet{huang2018hierarchically} leverage hierarchical decoding where a high-level decoder constructs a plan by generating a topic and a low-level decoder generates sentences based on the topic.
There have been a few works which try to incorporate real world knowledge during the process of story generation. \citet{guan2018story} use an incremental encoding (IE) scheme and perform one hop reasoning over the ConceptNet graph ConceptNet in order to augment the representation of words in the sentences. \citet{Chen2018IncorporatingSC} also tackle the problem in a similar way by including "commonsense knowledge" from ConceptNet as well. Several prior work focus on generating more coherent stories. \citet{clark2018neural} model entity representations explicitly by combining it with representations of the previous sentence and \citet{martin2018event} model events representations and then generate natural language sentences from those events (event2sentence). \citet{li-etal-2018-generating} use adversarial training to help the model generate more reasonable endings.

A common problem with such neural approaches in general is that the generated text is very "safe and boring". There has been a lot of recent efforts towards generating diverse outputs in problems such as dialogue systems, image captioning, story generation, etc., in order to alleviate the safe or boring text generation problem. Methods include using self-attention \citet{shao-etal-2017-generating}, Reinforcement Learning \cite{li-etal-2017-adversarial}, GANs etc.
\citet{xu2018diversity} proposed a method called Diversity-Promoting Generative Adversarial Network, which assigns low reward for repeatedly generated text and high reward for novel and fluent text using a language model based discriminator. 
\citet{li-etal-2016-diversity} propose a Maximum Mutual Information (MMI) objective function and show that this objective function leads to a decrease in the proportion of generic response sequences. \citet{nakamura2018another} propose another loss function for the same objective. In our models we experiment with their loss function and observe similar effects.

Recent works have also made advances in controllable generation of text based on constraints to make the outputs more specific. \citet{peng2018towards} have a conditional embedding matrix for valence to control the ending of the story. \citet{hu2017toward} have a toggle vector to introduce constraint on the output of text generation models using Variational Auto Encoders\cite{doersch2016tutorial}. Generating diverse responses based on conditioning has been done extensively in the field of dialogue systems. \citet{xing2016topic,zhou2018emotional, zhang2018personalizing} propose conditioning techniques by using emotion and persona while generating responses. Conditioned generation has also been studied in the field of story generation to plan and write \cite{yao2018plan, huang2018hierarchically} stories. 

In this work, we focus on generating more diverse and interesting endings for stories by introducing conditioning on keyphrases present in the story context and encouraging infrequent words in the outputs by modifying the training objective, thus leading to more interesting story endings.  

\section{Background}

\subsection{Sequence-to-Sequence with Attention}

We use a Seq2Seq model with attention as our baseline model. Words ($w_i$) belonging to the context of the story are fed one by one to the encoder (uni-directional but multi-layer) which produces the corresponding hidden representations $h^{enc}_{i}$. Finally, the hidden representation at the final timestep (T) is passed on to the decoder. During training, for each step t, the decoder (a unidirectional GRU) receives the word embedding of the previous word($x_{t-1}$) and the hidden state ($h^{dec}_{t-1}$). At training time, the word present in the target sentence at timestep $t-1$ is used and at test time, the actual word emitted by the decoder at the time step $t-1$ is used as input in the next time step.

However, to augment the hidden representation that is passed from encoder to the decoder, one can use the mechanism of attention \cite{bahdanau2014neural}. The attention weights at time step $t$ during decoding, denoted as $a^{t}$, can be calculated as:

\begin{align}
e^{t} &=v^{T} \tanh \left(W_{dec}\, h^{dec}_{t} + W_{enc} \, h^{enc}_{1:T_{src}} \right) \\
a^{t}_i &=\operatorname{softmax}\left(e^{t}_i\right)    
\end{align}

where $W_{dec}$, $W_{enc}$ and $v^{T}$ are learnable parameters and $a^{t}_i$ denotes the $i^{th}$ component of the attention weights. The attention weights can be viewed as a probability distribution over the source words, that tells the decoder where to look to produce the next word. Next, the attention weights are used to produce a weighted sum of the encoder hidden states, known as the context vector($c_{t}$): 

\begin{equation}
    c_{t}=\sum_{i=1}^{T_{src}} a_{i}^{t} h^{enc}_{i}
\end{equation}

This context vector is then concatenated with the embedding of the input word. It is used by the decoder to produce a probability distribution. $P^{\textnormal{vocab}}_{t}$ over the whole vocabulary:

\begin{equation}
\label{eqn:inference}
    P^{\textnormal{vocab}}_{t}= Decoder_{t}([x_{t-1}:c_t]; h_t)
\end{equation}

During training, the loss for timestep t is the negative log likelihood of the target word $w_{t}^{*}$ for that timestep. 

\begin{equation}
    \textnormal{loss}_{t} = -\log P \left( w_{t}^{*}\right)
\end{equation} 

Thus, the overall averaged loss of the whole sequence becomes: 

\begin{equation}
    \textnormal{loss}=\frac{1}{T_{dec}} \sum_{t=1}^{T_{dec}} \operatorname{loss}_{t}
\end{equation}

\section{Model Overview}


We now describe our model and its variations. We hypothesize that conditioning on the keyphrases present in the story context leads to more specific and interesting outputs. We experiment with several variants for incorporating keyphrases in our base model. We further adapt the loss suggested by \citet{nakamura2018another} to encourage the model to generate infrequent tokens.

\subsection{Keyphrase Conditioning}
In this section we briefly describe four different variants used for incorporating keyphrases from the story context.
We first extract top k keyphrases from the story context using the RAKE algorithm \cite{inbook}. RAKE determines the keyphrases and assigns scores in the text based on frequency and co-occurrences of words. We then sort the list of keyphrases by their corresponding scores and take the top-k of those. Note, each of the keyphrases can contain multiple words. Each of the word in a multi-word key phrase is assigned the same score as the keyphrase. 

We use the top-k keyphrases and ignore the rest. For that, we explicitly set the score of 0 to all the keyphrases which couldn't get to the top-k list. Next, the scores of these top-k keyphrases are normalized so that the total score sums to 1. We call this set keyphrases and its corresponding score vector ${p}$. ${p}$ is a vector with length equal to the length of the story context, and the value of every element $j$ is equal to score of the keyphrase to which the word $w_j$ belongs. 

In all the four model variants described next, we incorporate the score vector ${p}$ to encourage the model to condition on the keyphrases.

\subsubsection{Keyphrase Addition}
\label{KeywordAddition}
In a Seq2Seq model with attention, for every timestep \textit{t} during decoding, the model generates a distribution $a^t_i$ which is the weight given for a given source context word $w_{i}$. In this variant, the model is provided an additional keyphrase attention score vector ${p}$ along with the self-learnt attention weight vector $a^t$. To combine the two vectors, we simply add the values of the two vectors for each encoder position \textit{i} and normalize the final vector $a_{i}^{'t}$. 

\begin{align*}
    &a_{i}^{'t}=\sum_{i} a_{i}^{t} + p_{i} \\
    &a_{i}^{'t}= \frac{a_{i}^{'t}}{|a_{i}^{'t}|}
\end{align*}

Now at each time step $t$ of the decoder, we compute the new context vector as follows:
\begin{equation}
    c_{t}^{'}=\sum_{i=1}^{T_{src}} a_{i}^{'t} h^{enc}_{i}
\end{equation}

\subsubsection{Context Concatenation}
\label{ContextConcat}
This variation calculates two separate context vectors - one based on attention weights learnt by the model, and another based on the keyphrase attention score vector. Then both these context vector are concatenated. The intuition for this method comes from multi-head attention \cite{vaswani2017attention}, where different attention heads are used to compute attention on different parts of the encoder states. Similarly we also expect the model to capture salient features from both types of context vectors.
\begin{align}
        &k_{t} = \sum_{i} p_{i} h^{enc}_{i}\\
        &c'_{t}= [k_{t};c_{t}]
\end{align}
We use this new context vector to calculate our probabilities over the words as described in equation \ref{eqn:inference}.

\subsubsection{Coverage Loss}
\label{CoverageLoss}
This is a variant which implicitly encourages the model to pay attention to all words present in the context. We adapt the attention coverage based loss proposed by \citet{see-etal-2017-get}. It also helps in avoiding repeated attention across different timesteps while decoding. Due to this constraint, the model should focus on different words in the story and generate outputs conditioned on these words. The loss function which is presented in the paper is :
\begin{equation}
\operatorname{loss}_{t}=-\log P\left(w_{t}^{*}\right)+\lambda \sum_{i} \min \left(a_{i}^{t}, \mathit{s}_{i}\right)
\end{equation}
Here $\mathit{s}_{i}$ is the sum of attention weight vectors till \textit{i} time steps and $a_{i}^{t}$ is the attention weight vector for the current time step \textit{i}.

\subsubsection{Keyphrase Attention Loss}
\label{KeywordLoss}
In this variant, instead of explicitly forcing the model to attend on the keyphrases, we provide additional guidance to the model in order for it learn to attend to keyphrases. We introduce an attention similarity based loss. We first create a coverage vector $q$, which is the sum of the attention weights across all the decoders time steps. We then calculate Mean Squared Error loss between this vector and our keyphrase score vector $p$. This loss is calculated once per story after the whole decoding of the generated ending has finished. Unlike the first two variants, this method only nudges the model to pay more attention to the keyphrases instead of forcing attention on them. While backpropagating into the network, we use two losses. One is the original reconstruction loss which is used in Seq2Seq models and the other is this keyphrase based attention loss.
This can be summarised by the following set of equations.

\begin{equation}
    q = \sum_{t}^{T_{dec}} a^{t}
\end{equation}

\begin{equation}
     \textnormal{loss}_{\textit{keyphrase}} = \textnormal{MSE}(q, p)
\end{equation}
Where MSE is the mean squared error between the coverage vector and the probability distribution produced by the RAKE algorithm. This loss is weighted by a factor $\lambda$ and added to the cross-entropy loss.

\subsection{Inverse Token Frequency Loss}
As mentioned earlier, Seq2Seq models tend to generate frequent words and phrases, which lead to very generic story endings. This happens due to the use of conditional likelihood as the objective, especially in problems where there can be one-to-many correspondences between the input and outputs. MLE loss unfairly penalizes the model for generating rare words which would be correct candidates, but are not present in the ground truth. This holds for our problem setting too, where for the same story context, there can be multiple possible story endings. \citet{nakamura2018another} proposed an alternative Inverse Token Frequency (ITF) loss which assigns smaller loss for frequent token classes and larger loss for rare token classes during training. This encourages the model to generate rare words more frequently compared to cross-entropy loss and thus leads to more interesting story ending outputs.

\section{Experimental Setup}

\subsection{Dataset}
We used the ROCStories \cite{mostafazadeh2017lsdsem} corpus to generate our story endings. Each story in the dataset comprises of five sentences. The input is the first four sentences of the story and output is the last sentence of the story. The number of stories which were used to train and test the model are shown in Table \ref{table:dataset}.
\begin{table}[t!]
\begin{center}
\begin{tabular}{|l|l|l|l|}
\hline \bf  & \bf Train Set & \bf Dev set & \bf Test Set \\ \hline
ROC Stories & 90,000 & 4,081 & 4,081   \\
\hline
\end{tabular}
\end{center}
\caption{\label{table:dataset} The number of train, dev and  test stories in each of the ROCStories corpus.}
\end{table}

\subsection{Baselines and Proposed Methods}
For the evaluation of story ending generation, we compare the following baselines and proposed models:\\
\textbf{Seq2Seq:} Seq2Seq model with attention trained with vanilla Maximum likelihood Estimate(MLE) loss.\\
\textbf{IE + GA:} model based on Incremental Encoding (IE) and Graph Attention (GA) \cite{Guan2019StoryEG}. \\
\textbf{Seq2Seq + ITF:} Seq2Seq model with attention trained with ITF loss.\\
\textbf{Keyphrase Add + ITF:} Our model variant described in section \ref{KeywordAddition}. \\
\textbf{Context Concat + ITF:} Our model variant described in section \ref{ContextConcat}.\\
\textbf{Coverage Loss + ITF:} Our model variant described in section \ref{CoverageLoss} based on \cite{see-etal-2017-get}.\\
\textbf{Keyphrase Loss + ITF:} Our model variant described in section \ref{KeywordLoss}.\\
\textbf{Keyphrase Loss:} Our model variant described in section \ref{KeywordLoss} without the ITF loss.

\subsection{Experiment Settings}
All our models use the same hyper-parameters. We used a two layer encoder-decoder architecture with 512 GRU hidden units. We train our models using Adam optimizer with a learning rate of 0.001. For the Keyphrase Attention Loss model we assign the weight of 0.9 to Keyphrase loss and 0.1 to reconstruction loss. We use the best win percent from our Story-Cloze metric (described in the next section) for model selection. For ITF loss we use the hyperparameters mention in the original paper. We also apply basic heuristics to prevent continuous repetition of words.

\subsection{Automatic Evaluation Metrics}
In this section, we briefly describe the various metrics which were used to test our models. We did not use perplexity or BLEU as evaluation metric, as neither of them is likely to be an effective evaluation metric in our setting. This is since both these metrics measure performance based on a single reference story ending present in the test dataset, however there can be multiple valid possible story endings for a story. Therefore, we 

\textbf{DIST (Distinct):} 
Distinct-1,2,3 calculates numbers of distinct unigrams, bigrams and trigrams in the generated responses divided by the total numbers of unigrams, bigrams and trigrams. We denote the metrics as DIST-1,2,3 in the result tables. Higher Distinct scores indicate higher diversity in generated outputs.

\textbf{Story-Cloze:}
Since it is difficult to do human evaluation on all the stories, we use the Story-Cloze task \cite{mostafazadeh2017lsdsem} to create a metric in order to pick our best model and also to evaluate the efficacy of our model against Seq2Seq and its variants. This new proposed metric measures the semantic relevance of the generated ending with respect to the context.
In the Story-Cloze task, given two endings to a story the task is to pick the correct ending. We can use this task to identify the better of two endings. In order to do so, we fine-tune BERT \cite{devlin2018bert} to identify the true ending between two story candidates. The dataset for this task was obtained using the Story-Cloze task. Positive examples to BERT are obtained from the Story-Cloze dataset while the negative examples are obtained by randomly sampling from other story endings to get false ending for the story. We fine tune BERT in the two sentence setting by providing the context as the first sentence and the final sentence as the second. We pick the ending with a greater probability (from BERT's output head) of being a true ending as the winner. With this approach we were able to get a Story-Cloze test accuracy of 72\%. 

We now use this pre-trained model to compare the IE + GA model with our models. We select the winner based on the probability given by the pre-trained Bert model.

\subsection{Results}
We measure the performance of our models through automatic evaluation metrics as well as human evaluation. We use Distinct1, Distinct2 and Distinct3 to measure the diversity of our outputs. Additionally, we have built an automatic evaluation system using BERT and the Story-Cloze task following \citet{fan2018hierarchical} in order to compare our model against the state of the art models like the IE model. We also perform human evaluation on the stories generated by our model to get a overall sense of the model's performance.

\begin{table}[t]
\small
\begin{center}
\begin{tabular}{|l|l|l|l|}
\hline \bf Model & \bf DIST-1 & \bf DIST-2 & \bf DIST-3 \\ \hline
Seq2Seq & 0.039 & 0.165 &  0.335  \\
IE + GA & 0.026 &  0.130 &  0.263   \\
Seq2Seq + ITF & 0.063 & 0.281 & 0.517 \\
Keyphrase Add + ITF &  0.065 &  0.289&  0.539\\
Context Concat + ITF & 0.065 & 0.294& 0.558\\
Coverage Loss + ITF & 0.066 & 0.315& \textbf{0.590 } \\
Keyphrase Loss &  0.055 &  0.243 &  0.443\\
Keyphrase Loss + ITF & \textbf{0.068} & \textbf{0.318} & 0.588 \\
\hline
\end{tabular}
\end{center}
\caption{\label{table:models-metrics-distinct} Model comparison based on automatic metrics DIST-1, DIST-2 and DIST-3. }
\end{table}

\begin{table}[t]
\small
\begin{center}
\begin{tabular}{|l|l|l|l|}
\hline \bf Model & \bf Story-cloze \\ \hline
Seq2Seq & 49.8 \\
Seq2Seq + ITF & 54.1 \\
Keyphrase Add + ITF &  52.9\\
Context Concat  + ITF &  55.8\\
Coverage Loss + ITF  & 54.7 \\
Keyphrase Loss & 53.4\\
Keyphrase Loss + ITF &  \textbf{55.9}\\
\hline
\end{tabular}
\end{center}
\caption{\label{table:Story-Cloze} We measure the performance of the models using an automated Story-Cloze classifier which compares the outputs of model with the outputs of IE model.}
\end{table}

\begin{table}[t]
\small
\begin{center}
\begin{tabular}{|l|l|l|l|l@{}|}
\hline  \pbox{20cm}{\textbf{Number of} \\  \textbf{keyphrases}}   & \bf  DIST-1 & \bf DIST-2 & \bf DIST-3 & \textbf{Story-Cloze} \\ \hline
1 & 0.069 &  0.309 & 0.569  & 53.3\\
3  & 0.067 & 0.305 & 0.569 & \textbf{55.2} \\
5 & 0.066 & \textbf{ 0.315}& \textbf{0.590 } & 54.7\\
7 &  \textbf{0.072} & 0.315 &  0.575 & 53.9\\
All & 0.072 &  0.307 & 0.558 & 51.8\\

\hline
\end{tabular}
\end{center}
\caption{\label{table:metrics-distinct} Results for automatic metrics with varying number of keyphrases. Diversity is measured using DIST-1, DIST-2 and DIST-3 metrics. Story-Cloze loss measures relevance in comparison to IE model.}
\label{table:keyphrases}
\end{table}

\begin{table*}[htb]
\centering
\begin{tabular}{l|l|l} 
\hline
Example 1 & \makecell[l]{\textbf{Context} \\ \\  \\  \textbf{Seq2Seq Model}\\ \textbf{IE Model} \\ \textbf{Our Best Model} \\ \textbf{Ground Truth}} & \makecell[l]{megan was new to the pageant world . in fact , this was her very \\ first one . she was really enjoying herself , but was also quite nervous . \\ the results were in and she and the other contestants walked out . \\ she was disappointed that she couldn't have to learn how to win .\\ the next day , she was happy to have a new friend .\\ megan won the pageant competition . \\ they read off the names and megan had won ! }  \\
\hline
\hline
Example 2 & \makecell[l]{\textbf{Context} \\ \\  \\  \textbf{Seq2Seq Model}\\ \textbf{IE Model} \\ \textbf{Our Best Model} \\ \textbf{Ground Truth}} & \makecell[l]{alicia was going on a trip to disney . she was very excited .\\ yet when she got there she did not like it .\\ she thought it was too childish .\\ she had to go home .\\thankfully she got a call .\\ alicia decided to watch the movie instead . \\ she decided to never go back. } \\
\hline
\hline Example 3 & \makecell[l]{\textbf{Context} \\ \\  \\  \textbf{Seq2Seq Model}\\ \textbf{IE Model} \\ \textbf{Our Best Model} \\ \textbf{Ground Truth}} & \makecell[l]{katie found out her husband was cheating on her . she tried to \\ make  it work , but he wasn't interested . he took all her money \\ and divorced her . a few weeks later , his mistress dumped him \\katie was so happy , he helped her to help her out . \\ he was so upset , he decided to go home . \\ katie was devastated by himself and dumped her boyfriend .\\ she is much happier for being rid of him . }  \\
\hline
\hline
Example 4 & \makecell[l]{\textbf{Context} \\ \\  \\  \textbf{Seq2Seq Model}\\ \textbf{IE Model} \\ \textbf{Our Best Model} \\ \textbf{Ground Truth}} & \makecell[l]{tom had a cat . he noticed the litter wasn't being used .\\ after a while he started noticing a bad smell . he looked around \\ and  found a corner under the sofa .\\ he was able to get it back . 
\\he was able to get it out and clean it .
\\ tom cleaned the catnip smell up and cleaned it outside . \\it was full of cat poop that he had to clean up . 
 }
\end{tabular}
\caption{\label{table:qualitative} Qualitative analysis of the outputs produced by our model. For each context of the story, we show the endings generated by various models. It can be seen from the outputs that our model is able to generate specific outputs for a context.  }
\end{table*}

\subsubsection{Model Comparison and Ablation Study}
From the Table \ref{table:models-metrics-distinct}, we observe that the Seq2Seq model and the incremental encoding + graph attention (IE + GA) model have the worst performance in diversity. Although it has been shown that the IE + GA model achieves a good BLEU score, we observe that the model does not do as well on our automated metrics like DIST-1, 2 and 3 because the model has learnt to generate endings which match the distribution as a whole instead of generating story specific endings.

As expected, Seq2Seq + ITF loss model greatly outperforms the vanilla Seq2Seq model. As does the Keyphrase loss, showing that these models are indeed able to focus on different context words resulting in more diverse generations. 

The Story-Cloze based performance of the models is presented in Table \ref{table:Story-Cloze}.
The Keyphrase + ITF loss model outperforms all models on both the diversity and Story-Cloze metrics. Hence, we select Keyphrase + ITF loss model as the best model in further discussions. As an \textit{ablation} study, we run the Keyphrase loss model with the MLE loss instead of the ITF loss. We find that this model performs poorly than then its version with the ITF loss but still performs quite better than the Seq2Seq model. Also we note that the diversity obtained in Keyphrase + ITF loss model is greater than the Seq2Seq + ITF model and the Keyphrase loss model without ITF. It shows that a combination of both, Keyphrase attention loss and ITF loss, achieves better performance than these components by themselves.

\subsubsection{Effect of varying number of keyphrases}
In order to better understand the effect of keyphrases on the diversity and relevance of story endings, we ran the Coverage Loss model with varying number of keyphrases. Table \ref{table:keyphrases} shows the results of the experiment. We see that both Story-Cloze loss and DIST-1,2,3 are low when we use 1 keyphrase and also when we use all the keyphrases. This is expected, since in the case of 1 keyphrase, the model has very little keyphrase related information. In the other extreme case, providing all keyphrases covers a large proportion of the original context itself, and thus does not provide any extra benefit. We see good performance within the range of 3-5 keyphrases, where using 5 keyphrases gives the best diversity and 3 keyphrases gives the best Story-Cloze score. Informed by this experiment, we use 5 keyphrases in all our other experiments.

\subsection{Human Evaluation}
Since automatic metrics are not able to capture all qualitative aspects of the models, we performed a human evaluation study to compare our models. We first randomly selected 50 story contexts from the test set, and show them to three annotators. The annotators see the story context, and the story endings generated by our best model and the baseline IE+GA model in a random order. They are asked to select a better ending among the two based on three criteria - 1) \textit{Relevance} - Story ending should be appropriate and reasonable according to the story context. 2) \textit{Interestingness} - More interesting story ending should be preferred 3) \textit{Fluency} - Endings should be natural english and free of errors. We found that both models were preferred \textit{50\% }of the time, that is, both model was picked for 25 stories each. From a manual analysis of human evaluation, we found that our model was selected over the baseline in many cases for generating interesting endings, but was also equivalently penalized for losing the relevance in some of the story endings. We discuss this aspect in more detail in section \ref{sec:qa}.

\subsection{Qualitative Analysis}
\label{sec:qa}
In Table \ref{table:qualitative}, we show some example generations of our model and baselines. From example 1 and 2, it can be seen that the baseline models produce generic responses for story endings without focusing much on the context and keyphrases in the story. However, our model conditions on words like "pageant" in the story context, and includes it in the output even though it is a rare word in the corpus. Another point to note is that our model tends to include more proper nouns and entities in its output, like \emph{alicia} and \emph{megan} instead of using generic words like "he" and "she". However, our model is penalised a few times for being too adventurous, because it tends to generate more rare outputs based on the context. For example, in example 3, it got half of the output correct till "katie was devastated", but the other half "dumped her boyfriend" although is more interesting than the baseline models, is not relevant to the story context. The model incorrectly refers to katie with the pronoun "himself". In example 4, our model's generated output is quite relevant and interesting, apart from the token "catnip", for which it is penalized in human evaluation. Hence, although our model generates more interesting outputs, further work is needed to ensure that 1) The generated outputs entail the story context at both semantic and token level. 2) The generated output is logically sound and consistent. 

\section{Conclusion}
In this paper we have presented several models to overcome the generic responses produced by the state of the art story generation systems. We have both quantitatively and qualitatively shown that our model achieved meaningful improvements over the baselines.

\bibliography{acl2019}
\bibliographystyle{acl_natbib}

\appendix

\end{document}